\documentclass[times, review, 10pt]{elsarticle}

\usepackage{amssymb}
\usepackage{amsfonts}
\usepackage{amsmath}
\usepackage{dsfont}
\usepackage{mathrsfs}
\usepackage{textcomp}
\usepackage{gensymb} 
\usepackage{graphicx}
\usepackage{booktabs}
\usepackage{array}
\usepackage{pdflscape}
\usepackage{multirow}
\usepackage{color} 
\usepackage{multicol}
\usepackage{subcaption}
\usepackage{algorithm}
\usepackage{algorithmic}
\usepackage{hyperref}
\usepackage{cleveref}

\journal{.}

\graphicspath{{img}}

\newcommand{\rev}[1]{{\color{black}#1}}

\newcommand{\DmL}{D_m^{l}}
\newcommand{\DmR}{D_m^{r}}
\newcommand{\pleft}{\hat{p}_{l}}
\newcommand{\pright}{\hat{p}_{r}}

\begin{document}

\begin{frontmatter}

\title{Splitting criteria for ordinal decision trees: an experimental study}

\author[imibic,pduco]{Rafael Ayllón-Gavilán}
\author[uloyola]{Francisco José Martínez-Estudillo}
\author[uco]{David Guijo-Rubio\corref{cor}}\cortext[cor]{Department of Computer Science and Numerical Analysis, Universidad de Córdoba, Campus de Rabanales, Ctra. N-IVa, Km. 396, Córdoba, 14071.} \ead{dguijo@uco.es}
\author[uco]{César Hervás-Martínez}
\author[uco]{Pedro A. Gutiérrez}

\affiliation[imibic]{organization={Department of Clinical-Epidemiological Research in Primary Care, IMIBIC},
            addressline={Avda. Menéndez Pidal S/N},
            city={Córdoba},
            postcode={14004},
            country={Spain}}

\affiliation[pduco]{organization={Programa de doctorado en Computación Avanzada, Energía y Plasmas, Universidad de Córdoba},
            addressline={Campus de Rabanales, Ctra. N-IVa, Km. 396},
            city={Córdoba},
            postcode={14071},
            country={Spain}}
            
\affiliation[uloyola]{organization={Department of Quantitative Methods, Universidad Loyola Andalucía},
            addressline={C. Escritor Castilla Aguayo, 4},
            city={Córdoba},
            postcode={14004},
            country={Spain}}

\affiliation[uco]{organization={Departamento de Ciencia de la Computación e Inteligencia Artificial, Universidad de Córdoba},
            addressline={Campus de Rabanales, Ctra. N-IVa, Km. 396},
            city={Córdoba},
            postcode={14071},
            country={Spain}}

\begin{abstract}
Ordinal Classification (OC) addresses those classification tasks where the labels exhibit a natural order. Unlike nominal classification, which treats all classes as \rev{mutually exclusive and unordered}, OC takes the ordinal relationship into account, producing more accurate and relevant results. This is particularly critical in applications where the magnitude of classification errors \rev{has significant consequences}. Despite this, OC problems are often tackled using nominal methods, leading to suboptimal solutions. Although decision trees are \rev{among} the most popular classification approaches, ordinal tree-based approaches have received less attention when compared to other classifiers. This work provides a comprehensive survey of ordinal splitting criteria, standardising the notations used in the literature \rev{to enhance clarity and consistency}. Three ordinal splitting criteria, Ordinal Gini (OGini), Weighted Information Gain (WIG), and Ranking Impurity (RI), are compared to the nominal counterparts of the first two (Gini and information gain), by incorporating them into a decision tree classifier. An extensive repository considering $45$ publicly available OC datasets is presented, supporting the first experimental comparison of ordinal and nominal splitting criteria using well-known OC evaluation metrics. \rev{The results have been statistically analysed, highlighting that OGini stands out as the best ordinal splitting criterion to date, reducing the mean absolute error achieved by Gini by more than $3.02\%$.} \rev{To promote reproducibility, all source code developed, a detailed guide for reproducing the results, the $45$ OC datasets, and the individual results for all the evaluated methodologies are provided.}
\end{abstract}


\end{frontmatter}

\section{Introduction}
\rev{Ordinal Classification (OC) \cite{27_gutierrez2015ordinal} is a Machine Learning (ML) paradigm designed to address a specific type of problem in which class labels follow a natural ordering, but the exact distance between the labels is unknown. This field is also commonly referred to as ordinal regression. Traditionally, such problems have been tackled using standard (nominal) classification techniques \cite{large2018detecting,guijo2021statistical}. However, nominal classifiers disregard the intrinsic ordering of the labels, often leading to suboptimal solutions. In these cases, OC is better suited to handle the problem, as it explicitly incorporates the ordinal nature of the target variable into the learning process.}

For example, in computed tomographic images of lung nodules \cite{armato2011lung}, each image is assigned a label ranging from $1$ to $5$ representing the malignancy score. A value of $1$ indicates a low likelihood of malignancy, $3$ represents indeterminate malignancy, and $5$ reflects a high probability of malignancy. It is clear that classifying an image as indeterminate (scores $2$ or $3$) or even as non-malignant (score $1$), when it is actually highly malignant (score $5$), should be more penalised than classifying a non-malignant image (score $1$) as indeterminate (scores $2$ or $3$). This example perfectly reflects that the primary objective of OC is to bear in mind this inherent ordering to develop more accurate models.

\begin{figure}[ht!]
    \centering
    \includegraphics[width=0.75\linewidth]{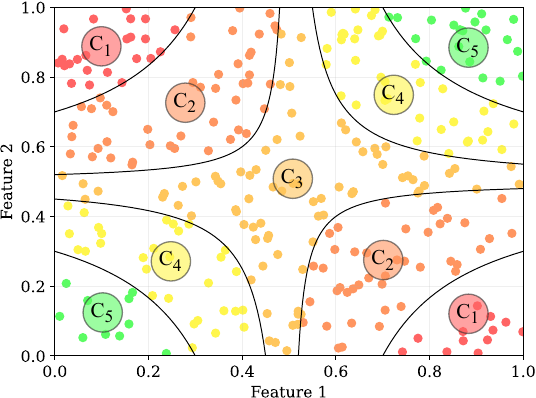}
    \caption{\rev{Synthetic OC problem where samples are represented in a 2D feature space. These samples are coloured according to the class they belong. Ordinality is expressed according to distance from centre in the feature space. More information about this problem can be found in \cite{cardoso2010classification}.}}
    \label{fig:ordinal-example}
\end{figure}

A synthetic example of an OC problem is illustrated in \Cref{fig:ordinal-example}, where samples are represented in a 2D feature space. Each sample is coloured according to its assigned class, with five different classes labelled from $C_1$ (red) to $C_5$ (green). The ordinal nature of the problem is evident, as the distance in the feature space increases with the ordinal difference between the classes. The boundaries separating the classes are shown, highlighting the gradual transition between adjacent classes and the greater distinction between classes further apart on the ordinal scale.

OC problems can be found in a wide variety of fields, including\rev{, as previously noted,} medical research \cite{duran2021ordinal,lei2022meta}, automation science and engineering \cite{09_xian2018causation, tang2024catnet}, wave height prediction \cite{gomez2024orfeo}, credit rating \cite{goldmann2024new}, face recognition \cite{he2018learning, xu2024facial}, age estimation \cite{vargas2024age}, image classification \cite{17_zhu2021convolutional}, wind speed prediction \cite{pelaez2024general}, manufacturing \cite{19_wang2021objective}, text classification \cite{21_baccianella2014feature}, and transportation \cite{22_yildirim2019eboc}, among others. These studies exemplify the use of specifically designed OC models, \rev{which exploit} ordinal information to \rev{outperform} their nominal counterparts.

The OC paradigm has considerably evolved in recent years, thanks to substantial advances in supervised learning methodologies \cite{23_seah2012transductive}. This includes contributions ranging from neural network models, like the one proposed in \cite{lazaro2023neural}, which also addresses the imbalance problem typical of ordinal tasks by minimizing a Bayesian cost, to deep ordinal models \cite{kook2022deep}, soft-labelling methods \cite{vargas2022unimodal, vargas2023generalised} and ordinal representation learning techniques \cite{LEI2024110748}. Tree-based models for OC \rev{have received increasing attention} in the last decade \cite{28_tutz2022ordinal,42_marudi2024decision}. One of the primary advantages of tree-based approaches is their ability to produce interpretable models that facilitate the understanding of complex and non monotonic dependencies between explanatory attributes and the dependent variable, also referred \rev{to} as output variable. For instance, the simplest tree-based models, the Decision Trees (DTs), can generate easy-to-interpret rules without compromising performance. Another strength of DTs is their versatility, as they can handle continuous, discrete and categorical inputs while being invariant under strictly monotone transformations of individual inputs. In addition, an internal feature selection is performed as an integral part of the training process.

The idea behind a DT involves recursively partitioning the predictor space, also known as input space, into a set of rectangles according to a splitting criterion. Subsequently, a simple model, such as a \rev{constant prediction}, is fitted to each rectangle. In standard nominal classification, the most widely used approaches are CART \cite{31_leo1984classification}, along with other variants such as ID3 \cite{33_quinlan1986induction} and C4.5 \cite{32_quinlan2014c4}. 

The splitting criterion plays a key role in the learning process, as it determines how the predictor space is partitioned at each step, thereby guiding the construction of the tree. The splitting criteria used to assess the quality of a split is based on the concept of impurity, which quantifies the homogeneity of the labels at a given node. The objective during splitting is to minimise the impurity. Consequently, there is a strong relationship between designing effective splitting criteria and developing appropriate impurity measures. In nominal classification, the Gini-index \cite{31_leo1984classification} and Shannon-entropy \cite{shannon1948mathematical} are the most used impurity measures, forming the basis of the well-known Gini and Information Gain (IG) splitting criteria, respectively. \rev{In this way, splitting criteria can be applied to theoretical and practical domains. For instance, in \cite{mamdouh2022new}, splitting criteria are utilised to perform feature selection in the context of text classification. In \cite{jiao2023dtec}, decision trees with different splitting criteria are employed to discover interpretable partitions of uncertain data. In \cite{mostafa2024feature}, splitting criteria are employed to perform feature reduction in hepatocellular carcinoma prediction.}


Specifically, the Gini-index impurity measure computes the expected loss on all classes due to misclassification errors, whereas the Shannon-entropy represents the uncertainty of a random variable outcome. However, both Gini-index and Shannon-entropy, to be minimised, do not take into account the ordinal properties of the dependent variable. For instance, \rev{suppose} the following two nodes representing the distribution of $20$ observations into $4$ ordered classes: A) $(10,0,0,10)$, and B) $(10,10,0,0)$. Considering these two distributions, we can observe that both Gini-index and Shannon-entropy \rev{yield} the same value ($0.5$ and $1.0$, respectively). However, from an ordinal perspective, option B is preferable over option A, \rev{as it better separates the classes along the ordinal scale}. This example is graphically illustrated in \Cref{fig:impurity_measures}, where the values for ordinal impurity measures are also presented \rev{and discussed in more detail, together with their associated splitting criteria, in \Cref{sec:methodology}}. This observation extends to the splitting criteria, to be maximised, defined by these impurity measures. \Cref{fig:splitting_criteria} illustrates two different splits of a given node, for which the nominal splitting criteria \rev{(Gini and IG)} achieve the same value regardless the ordinality considered in the sub-nodes. \rev{Therefore, nominal splitting criteria are suboptimal in ordinal tasks, as the ordering information of the classes is not leveraged. The choice of an appropriate splitting criterion is crucial for improving the performance of tree-based techniques in practical applications, as both the partitioning of the input space and the embedded feature selection during training are highly dependent on the splitting criterion employed.}

\rev{To address this limitation, this work focuses on the analysis of splitting criteria utilised in tree-based methods, specifically designed to incorporate and exploit the ordinal nature of the data. These methods aim to enhance predictive performance while preserving the inherent structure of ordinal relationships. Therefore, this work is guided by the following research question: \textit{Could the use of ordinal splitting criteria in tree-based methods improve the performance of their nominal counterparts?}}

\rev{The specific objectives that have been defined to address this research question are as follows:
\begin{enumerate}
    \item To study and analyse the ordinal splitting criteria proposed in the literature.
    \item To conduct a comprehensive comparison between ordinal splitting criteria and their nominal counterparts.
    \item To evaluate the performance of different ordinal splitting criteria across a variety of ordinal classification problems.
\end{enumerate}}

\rev{Note that these splitting criteria are evaluated in the context of decision tree classifiers, the application of these criteria in tree ensemble contexts such as Random Forests or Boosting being beyond the scope of this work}.

The rest of the paper is organised as follows: \rev{\Cref{sec:contributions} details the main key contributions of this work}; \Cref{sec:lit_review} reviews commonly used impurity measures (\Cref{subsec:impurity}) and main OC tree-based approaches (\Cref{subsec:OCtrees}); \Cref{sec:methodology} presents the nominal and ordinal impurity measures and their corresponding splitting criteria; \Cref{sec:experimental} describes the experimental study conducted; the results achieved are included and statistically analysed in \Cref{sec:results}; \rev{and finally, this work is closed with some final remarks (\Cref{sec:conclusions}), and limitations and future research directions (\Cref{sec:future})}.

\color{black}
\section{Key contributions} \label{sec:contributions}

In this work, we present a thorough theoretical study and empirical evaluation of three ordinal splitting criteria for tree-based methods in the context of OC. This analysis includes a detailed comparison with two conventional (nominal) splitting approaches, which highlights their limitations when applied to OC problems. The validity of these ordinal splitting criteria has been demonstrated through their application to a diverse set of $45$ OC problems. The results demonstrate the potential of ordinal splitting methods to enhance predictive performance.

The key contributions of this paper are summarised as follows:
\begin{enumerate}
    \item To provide a comprehensive description and survey of existing splitting criteria for tree-based approaches in OC.
    \item To standardise and unify the diverse notations previously used in the literature for existing ordinal splitting criteria, enabling consistency and clarity.
    \item To present an extensive OC repository including a total of $45$ publicly available datasets to facilitate research and benchmarking.
    \item To perform the first experimental study comparing the performance of several splitting criteria with their nominal counterparts. This comparison is conducted using three well-known performance metrics: Mean Absolute Error (MAE), Quadratic Weighted Kappa (QWK), and Ranked Probability Score (RPS).
    \item To provide practitioners with a comprehensive \texttt{Python} implementation of various splitting criteria and the DT algorithm\footnote{\url{https://github.com/ayrna/decision-trees-from-scratch}}. We have developed a new framework that serves as a versatile platform for testing and validating novel approaches and splitting criteria in a plug-and-play way. Furthermore, the OC dataset repository and the experimental results are made available on the associated webpage\footnote{\url{https://www.uco.es/grupos/ayrna/ordinal-trees}}.
\end{enumerate}
\color{black}

\section{Literature review} \label{sec:lit_review}

This section examines the most common impurity measures used in DT for nominal classification, discussing their limitations in capturing ordinal relationships and highlighting the need for more appropriate impurity measures for OC problems. Then, a review of the main tree-based methodologies for OC is provided, addressing a gap in the literature as no prior surveys have specifically explored this area. \rev{For instance, in \cite{27_gutierrez2015ordinal} and in \cite{28_tutz2022ordinal}, two ordinal classification surveys published in 2016 and 2022, respectively, no ordinal splitting criteria are discussed.}

\subsection{Current nominal and ordinal impurity measures: limitations and scope} \label{subsec:impurity}

An impurity measure is a metric that estimates the probability of misclassifying a randomly chosen sample from a labelled set. This \rev{can also be interpreted as a measure of the heterogeneity of a labelled dataset in terms of class distribution}. A higher impurity value indicates greater heterogeneity, while a lower value reflects greater homogeneity. The Gini-index and Shannon-entropy \rev{are the two most commonly used measures in this context}.

A generalisation of the Gini-index impurity specifically designed for DT-based methodologies was proposed in \cite{31_leo1984classification}, \rev{which is expressed} as the misclassification cost of assigning a sample to \rev{an incorrect} class. An extension of the Gini-index appropriate for OC was presented in \cite{35_piccarreta2008classification}, known as Ordinal Gini-index (OGini-index). The idea behind OGini-index is to employ the cumulative frequency of each class instead of the relative frequency to measure the heterogeneity. 

On the other hand, the Shannon-entropy ($H$) \cite{shannon1948mathematical} is \rev{defined} based on the probabilities of each class. Several works have extended the Shannon-entropy to \rev{take into account the ordinal nature of the classes}. However, most of them assume monotonicity not only of the output but also of the input variables \cite{37_potharst2002classification, 38_hu2011rank}. Recently, a variant of Shannon-entropy was introduced in \cite{41_kelbert2017weighted}, referred to as Weighted Entropy ($H_w$). Unlike traditional entropy measures, this impurity measure does not depend on the assumption of \rev{monotonicity} and was specifically adapted for ordinal DTs in \cite{40_singer2020weighted}. The weighting mechanism in this work was designed to account for the ordinal nature of the dataset by calculating the distances between classes relative to the mode class, which represents the most probable class. Furthermore, $H_w$ was successfully applied to Random Forest (RF) and AdaBoost algorithms \cite{singer2020ordinal, 39_singer2020objective}, extending its utility to ensemble methods. Building on these advancements, \cite{42_marudi2024decision} provided a generalisation of $H_w$, making it applicable to any DT-based methodology.

\rev{Another impurity measure is the Ranking Impurity (RI)} \cite{34_xia2006effective}, which, unlike previous impurity measures, was not derived from a nominal counterpart. The RI can be interpreted as the maximum potential number of \rev{misranked} pairs in the set. Here, the miss-ranked pairs are weighted by the difference between the assigned scores.

\rev{As can be observed}, most of the aforementioned ordinal impurity measures depend on the assignment of numerical scores to the classes, thereby using a higher scale level. Specifically, \rev{$H_w$ and RI} explicitly incorporate these assigned scores into their definitions. In contrast, OGini-index is score-free as its definition is based on the cumulative frequency of each class. While the assignment of scores may be justified in certain scenarios, particularly when ordinal labels are built from continuous variables by discretisation, it is rather artificial and arbitrary in the context of genuine ordinal data. For instance, when the response variable represents ordered levels of disease severity, assigning numerical scores might not align with the natural characteristics of the data. \rev{In this work, we use the three aforementioned ordinal impurity measures to define distinct splitting criteria}, which are thoroughly defined and mathematically formalised in \Cref{sec:methodology}.

\subsection{Existing ordinal classification trees} \label{subsec:OCtrees}

The first review on ordinal classification, to the best of the authors' knowledge, was presented in \cite{27_gutierrez2015ordinal}, where a general taxonomy was proposed based on \rev{the way in which models exploit ordinal information}. In this survey, the approaches were categorised into three distinct groups: 1) \rev{naïve} approaches, which involve standard ML algorithms designed for nominal classification or regression; 2) ordinal binary decomposition techniques, where an ordinal problem with $Q$ classes is decomposed into $Q$ binary classification problems; and 3) threshold-based techniques, which assume an underlying continuous variable that is modelled using a set of thresholds that map intervals on the real line to the corresponding ordinal labels.

\rev{An updated review was presented in} \cite{28_tutz2022ordinal} with an up-to-date taxonomy covering novel approaches and most prominent methodologies. In this case, the taxonomy includes the following 4 groups: 1) basic OC models, which include cumulative models, sequential models or adjacent \rev{category} models, among others; 2) models with more complex parametrisations, which extend previous basic models by including category-specific effects or \rev{capturing} intricate relationships between inputs; 3) hierarchically structured models, \rev{which} are flexible approaches for ordinal data that exploit the inherent structure of response categories by sequentially modelling grouped and subdivided categories; and 4) finite mixture models, which combine content-driven labels with uncertainty or response patterns, using probabilistic mixtures to capture heterogeneity.

\rev{However, some approaches do not fit neatly into either taxonomy}. \rev{Tree-based methods are an example, with some proposals in recent years.} Beyond the approaches detailed in \Cref{subsec:impurity}, additional methods have been introduced in the literature, such as the \rev{method} proposed in \cite{cardoso2010classification}, which is based on the concept of consistency, a global property that characterises the relationship between different decision regions within the input space. This property was integrated into the design of the learning algorithm \rev{to guide} the node expansion process to ensure alignment with the ordinal structure of the data.

Furthermore, building upon the RF methodology, three notable approaches have been introduced in the literature. The first one, known as Ordinal Forest \cite{30_hornung2018ordinalforest}, leverages the latent variable underlying the observed ordinal output. This method combines the strengths of ensemble models with the threshold-based approach to reduce the bias and variance associated with individual ordinal classifiers. The second study, presented by Janitza \textit{et al.} in \cite{48_janitza2016random}, extends RF by incorporating conditional inference trees, offering a robust framework for OC tasks. Finally, the third \rev{approach introduces} a variant of ordinal trees (also used in RF) based on binary models that are implicitly used in parametric OC \cite{tutz2022ordinal_2}.

\section{Methodology} \label{sec:methodology}

\rev{Let us} define a dataset $D$ with $N$ patterns as $D = \{(\mathbf{x}_1, y_1), \ldots, (\mathbf{x}_i, y_i), \ldots, (\mathbf{x}_N, y_N)\}$ where $\mathbf{x}_i \in \mathcal{X} \subseteq \mathbb{R}^K $ represents a vector of $K$ independent variables (or features), and $y_i \in \mathcal{Y} = \{C_1, C_2, \ldots, C_q, \ldots, C_Q\}$ is the label assigned to input pattern $\mathbf{x}_i$, i.e. the dependent variable. Since this work focuses on ordinal classification problems, it is assumed that $C_1 \prec C_2 \prec \ldots \prec C_Q $, where $\prec$ represents the order relationship. \rev{A common} approach in ordinal problems is to assign a numerical value to each category $C_q \in \mathcal{Y}$. This assignment is defined by a function $v$, \rev{typically defined as} $v(C_q) = q$, satisfying that $v(C_1) < v(C_2) < \ldots < v(C_q) < \ldots < v(C_Q)$. An analogy to this ordering can be found in regression problems, where the $\prec$ is replaced by the standard $<$ operator, and $\mathcal{Y} = \mathbb{R}$. However, unlike regression, in ordinal classification, the exact distance between classes is unknown. Nevertheless, it is assumed that different penalisation should be given to different classification errors, such that misclassifying a pattern into an adjacent class incurs a smaller penalty compared to misclassifying it into a class farther away on the ordinal scale.

In general terms, a DT is constructed by recursively partitioning the input dataset, so that patterns belonging to the same class are grouped together and separated as much as possible from patterns belonging to different classes. At each node $m$ of a DT, we work with a subset $D_m \subseteq D$. For the root node ($m=0$), $D_m = D$. During the subsequent steps of the growth phase, the goal \rev{at node $m$}, is to partition $D_m$ into two subsets $\DmL$ (left sub-node) and $\DmR$ (right sub-node). This partition is made by a split $\theta(k, t_k)$, where $k$ denotes one of the $K$ features, and $t_k$, is a real-valued threshold that specifies the splitting point for the feature $k$. Then, the two subsets $\DmL$ and $\DmR$ are defined as follows:
\begin{equation}
    \begin{aligned}
        \DmL = \{(\mathbf{x}_i, y_i) \mid x_{i_k} < t_k\}, \\
        \DmR = \{(\mathbf{x}_i, y_i) \mid x_{i_k} \geq t_k\},
    \end{aligned}
    \label{d_left_right}
\end{equation}
where $x_{i_k}$ is the value for the $k$-th feature of the $i$-th pattern. According to \cite{31_leo1984classification}, the optimal split $\theta(k^*, t_k^*)$ is the one maximising the real-valued function $\phi(D_m, \theta) \in \mathbb{R}$, which is commonly known as the splitting criterion. In general, the goal of the $\phi$ function is to quantify the impurity decrease that results from splitting a node, hence, it has to be maximised. \rev{As previously defined}, the impurity is a measure of the heterogeneity with respect to the output labels $\mathcal{Y}_m = \{y_i \in \mathcal{Y}: y_i \in D_m\}$ within a node $m$. Since the development of DTs, numerous splitting criteria have been introduced in the literature. Among these, the Gini and Information Gain (IG), derived from the Gini-index and Shannon-Entropy impurity measures, respectively, have demonstrated the strongest performance \cite{buntine1992further}.

\subsection{Gini}

The Gini-index is defined as a function $G$ that quantifies the probability of misclassifying a \rev{randomly drawn sample from the node}, assuming it was labelled based on the class distribution of that node. The Gini-index of a given set $D_m$ is defined as:
\begin{equation}
    G(D_m) = \hat{p}_{q|m} \left( 1 - \hat{p}_{q|m}\right) = 1 - \sum_{q=1}^{Q}{\left( \hat{p}_{q|m}\right)^2},
    \label{eq:gini}
\end{equation}
where $\hat{p}_{q|m} = \frac{N_q(D_m)}{N(D_m)}$ is the relative frequency of class $C_q$ at the $m$ node,  $N_q(D_m)$ represents the number of patterns belonging to category $C_q$ in $D_m$, and $N(D_m)$ is the total number of patterns in $D_m$.

Therefore, the splitting criterion based on the Gini-index ($\phi_{Gini}$) measures the impurity decrease caused by splitting a node $m$ into its \rev{left and right children:}
\begin{equation}
    \phi_{\text{Gini}}(D_m, \theta) = G(D_m) - ( \pleft \ G(\DmL) + \pright \ G(\DmR)),
    \label{eq:sp_gini}
\end{equation}
where $\pleft = \frac{N(\DmL)}{N(D_m)}$ and $\pright = \frac{N(\DmR)}{N(D_m)}$ are the proportion of samples in the left and right nodes, respectively. \rev{For simplicity, $\phi_{Gini}$ will be referred to as Gini}.

The Gini splitting criterion is highly effective for standard classification tasks. However, it disregards the ordinal relationship among classes. As a result, misclassifying a sample always incurs the same amount of loss, ignoring the distance in the ordinal scale between the observed and the predicted classes.

\subsection{Ordinal Gini (OGini)}

For ordinal classification, a variant of the Gini-index impurity measure, known as Ordinal Gini-index (OGini-index), was proposed in \cite{35_piccarreta2008classification}, defined as a function $OG$. The idea was to measure the heterogeneity between the two sub-nodes using the dissimilarity between the cumulative frequencies:
\begin{equation}
    OG(D_m) = \sum_{q=1}^{Q}\hat{c}_{q|m} (1 - \hat{c}_{q|m}),
    \label{eq:og}
\end{equation}
where $\hat{c}_{q|m} = \sum_{j=1}^{q}p_{j|m}$ is the cumulative frequency of class $C_q$ in $D_m$. Then, the splitting criterion based on $OG$ ($\phi_{OGini}$, for simplicity, denoted as OGini) is expressed as follows:
\begin{equation}
    \phi_{\text{OGini}}(D_m, \theta) = OG(D_m) - ( \pleft \ OG(\DmL) + \pright \ OG(\DmR)).
    \label{eq:sp_ogini}
\end{equation}

\subsection{Information Gain (IG)}

The IG splitting criterion is based on the Shannon-entropy impurity measure \cite{shannon1948mathematical}. In a DT setup, entropy, $H$, quantifies the uncertainty or disorder in a given node, $m$. If the node is perfectly homogeneous (all instances belong to the same class), $H=0$ because there is no uncertainty. On the other hand, if the node is equally split between classes, $H$ is at its maximum. Hence, $H$ at a node $m$ is defined as follows:
\begin{equation}
    H(D_m) = - \sum_{q = 1}^{Q}\left[\hat{p}_{q|m} \ log(\hat{p}_{q|m})\right].
    \label{eq:ig}
\end{equation}

Therefore, the IG splitting criterion, $\phi_{IG}$, quantifies the reduction in $H$ caused by splitting a node as follows:
\begin{equation}
\phi_{\text{IG}}(D_m, \theta) = H(D_m) - ( \pleft \ H(\DmL) + \pright \ H(\DmR)).
\label{eq:sp_ig}
\end{equation}

\subsection{Weighted Information Gain (WIG)}

The standard entropy measure, $H$, does not take into account the ordinal properties of the target variable $\mathcal{Y}$. Different variants of $H$ were proposed in \cite{41_kelbert2017weighted}, standing out the weighted entropy, $H_w$. This approach allocates weights to the different ordinal classes in such a way that the weight $w_q(D)$, assigned to the class $C_q$, is derived from the difference between the value of the class and the value of the mode class, i.e. the most probable class. This weight is defined as:
\begin{equation}
    w_q(D) = \frac{|v(C_q) - v(C_{\text{mode}})|^\propto}{\sum^{Q}_{j=1}|v(C_j) - v(C_{\text{mode}})|^\propto},
\end{equation}
where $C_{\text{mode}}$ is the mode class, i.e. the class with more instances in $D$, and $\propto$ is a normalisation parameter that balances the distribution of weights over the different classes. The introduction of weights is intended to consider the distribution of classes along the ordinal scale. Therefore, $H_w$, is defined as:
\begin{equation}
    H_w(D_m) = - \sum_{q = 1}^{Q} \left[w_q(D_m) \ \hat{p}_{q|m} \ \log(\hat{p}_{q|m})\right].
    \label{eq:hw}
\end{equation}

With the goal of considering the ordinal properties of the ordinal output, the Weighted Information Gain (WIG) splitting criterion, $\phi_{WIG}$, was introduced in \cite{40_singer2020weighted}, replacing $H$ of \Cref{eq:IG} by its weighted version, $H_w$, giving birth to:
\begin{equation}
    \phi_{\text{WIG}}(D_m, \theta) = H_w(D_m) - ( \pleft \ H_w(\DmL) + \pright \ H_w(\DmR)).
    \label{eq:sp_wig}
\end{equation}

It is important to highlight that the computation of the weights $w_q(D)$ for each class depends on the scores assigned to the ordered categories of the response, i.e. the function $v$ under consideration.

Additionally, different variations of weighted entropy, $H_w$, can be found in \cite{singer2020ordinal, 39_singer2020objective}, replacing, for instance, the mode of the classes by the highest class value, or employing those weights in an entropy-based objective function used to build the tree.

\subsection{Ranking Impurity (RI)}

The Ranking Impurity (RI) was presented in \cite{rankingImpurityXia2006}, with the focus of measuring the potential maximum number of miss-classified samples based on the current node distribution. A higher impurity value indicates more potential for misalignment between the observed and predicted classes. The RI at node $m$ is defined as follows:
\begin{equation}
    RI(D_m) = \sum_{q=1}^Q \sum_{j=1}^q \left[\beta(C_q, C_j) \ N_q(D_m) \ N_j(D_m)\right],
    \label{eq:ri}
\end{equation}
where $\beta$ is a weighting function that penalises the impurity according to the ordinal scale. The introduction of $\beta$ generalises the expression presented in \cite{rankingImpurityXia2006}, the one considered in our experimentation, in which $\beta(C_q, C_j) = v(C_q) - v(C_j)$. The splitting criterion based on RI ($\phi_{RI}$, for simplicity, the splitting criterion also adopts the RI acronym) is defined as:
\begin{equation}
    \phi_{\text{RI}}(D_m, \theta) = RI(D_m) - ( \pleft \ RI(\DmL) + \pright \ RI(\DmR)).
    \label{eq:sp_ri}
\end{equation}

In summary, the Gini and OGini splitting criteria are based on the Gini-index and OGini-index impurity measures, respectively. The standard IG and its ordinal counterpart, WIG, are derived from the standard entropy, $H$, and its weighted variant, $H_w$, respectively. Then, the RI splitting criterion is directly associated with the RI impurity measure.

\begin{figure}[ht]
	\centering
	\includegraphics[width=0.75\linewidth]{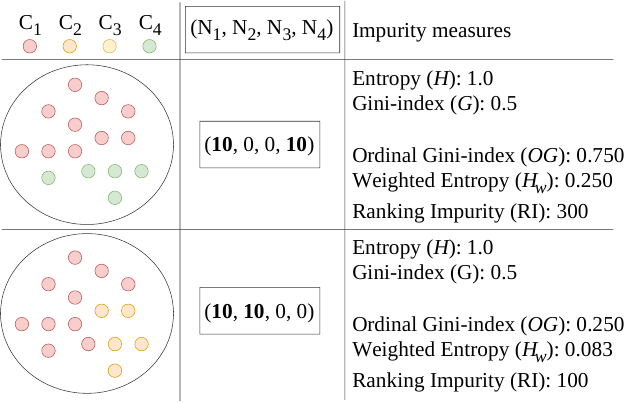}
	\caption{\rev{Representation of two different ordinal distributions for a problem with four classes, and associated values for the different impurity measures considered in this work.}}
	\label{fig:impurity_measures}
\end{figure}

\begin{figure}[ht]
	\centering
	\includegraphics[width=0.75\linewidth]{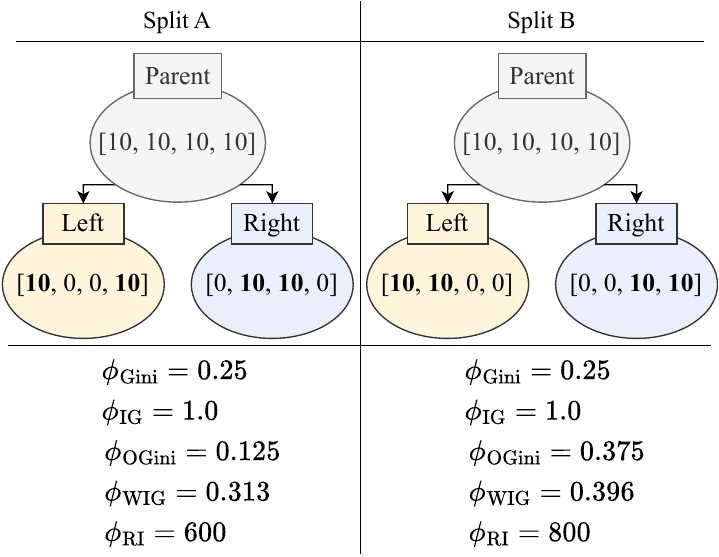}
	\caption{\rev{Comparison of two node splits based on nominal and ordinal splitting criteria. The split A does not preserve the ordinal relationship among classes, as first and last classes are grouped in the same node (left sub-node). Conversely, the split B maintains the ordinal structure, given that contiguous classes are grouped in the same sub-node (first and second classes are in the left sub-node, whereas third and fourth classes are in the right sub-node). Then, higher values for ordinal splitting criteria ($\phi_{\text{OGini}}$, $\phi_{\text{WIG}}$, and $\phi_{\text{RI}}$) are achieved. Note that nominal splitting criteria ($\phi_{\text{Gini}}$ and $\phi_{\text{IG}}$) remain unaffected by the ordinality of the splits.}}
	\label{fig:splitting_criteria}
\end{figure}

The way impurity measures and splitting criteria are computed is graphically illustrated in \Cref{fig:impurity_measures,fig:splitting_criteria}. Regarding impurity measures, \Cref{fig:impurity_measures} depicts two distinct ordinal distributions for a problem with four classes. Note that impurity measures are to be minimised. From the perspective of an ordinal scale, the upper distribution exhibits poorer separation, as distant classes are grouped together. Conversely, the lower distribution demonstrates better separation by grouping adjacent classes. Nominal impurity measures (Gini and standard entropy, $H$) fail to account for this distinction. However, the ordinal impurity measures ($OG$, weighted entropy -- $H_w$, and $RI$) explicitly consider these differences and consequently assign more favourable (lower) values to the distribution located below.

\rev{\Cref{fig:splitting_criteria} illustrates two different node splits (A and B), highlighting the differences between ordinal (OGini, WIG, and RI) and nominal splitting criteria (Gini and IG). The values in the figure represent splitting criteria that are maximised during the growing procedure. Specifically, the split A divides the patterns such that the left sub-node contains all patterns for $C_1$ and $C_4$, while the right sub-node contains patterns for $C_2$ and $C_3$. In this case, the left sub-node does not maintain an ordinal arrangement, as the classes within it are not contiguous. Conversely, the split B adheres to an ordinal structure: the left sub-node includes patterns for $C_1$ and $C_2$, and the right sub-node contains patterns for $C_3$ and $C_4$, preserving the order relationship. This distinction is also reflected in the splitting criteria values presented below the nodes, where higher values are observed for the ordinal splitting criteria in the right split. Notably, nominal splitting criteria achieve the same values regardless of whether the splits maintain ordinality or not.}

\rev{Hereafter, in order to enhance readability, we will refer to each splitting criterion by the sub-index of the $\phi$ function, e.g. we will refer to $\phi_{\text{OGini}}$ as OGini. In \Cref{tab:summary_sp}, we present a summary of the splitting criteria presented in this section, indicating the acronym, impurity measure, whether the splitting criterion is score-free or not, i.e. requires a mapping function such as $v$, the hyperparameters of the criterion, and the literature reference.}

\begin{table}[]
\color{black}
    \centering
    \setlength{\tabcolsep}{2.5pt}
    \begin{tabular}{clcccc}
         \toprule\toprule
            & Splitting criteria acronym      & Impurity             & Score-free    & Hyperparam.            & Ref. \\
            \midrule
            \multirow{2}{*}{\rotatebox[origin=c]{90}{Nominal}} & Gini ($\phi_{\text{Gini}}$, \Cref{eq:sp_gini})        & $G$ (\Cref{eq:gini}) & Yes           & No                     & \cite{31_leo1984classification} \\
            & IG ($\phi_{\text{IG}}$, \Cref{eq:sp_ig})         & $H$ (\Cref{eq:ig})   & Yes           & No                     & \cite{31_leo1984classification} \\
            \midrule
            \multirow{3}{*}{\rotatebox[origin=c]{90}{Ordinal}} & OGini ($\phi_{\text{OGini}}$, \Cref{eq:sp_ogini})      & $OG$ (\Cref{eq:og})  & Yes           & No                     & \cite{ordinalGiniPiccarreta2008} \\
            & WIG ($\phi_{\text{WIG}}$, \Cref{eq:sp_wig})        & $WIG$ (\Cref{eq:hw}) & No            & \{$\propto$, $v$\}     & \cite{40_singer2020weighted} \\
            & RI ($\phi_{\text{RI}}$, \Cref{eq:sp_ri})         & $RI$ (\Cref{eq:ri})  & No            & \{$\beta$, $v$\}       & \cite{rankingImpurityXia2006} \\
            \bottomrule\bottomrule
    \end{tabular}
    \caption{\rev{Summary of the splitting criteria discussed in this work.}}
    \label{tab:summary_sp}
\end{table}

\section{Experimental setup}\label{sec:experimental}

This section describes the experimental framework used to evaluate the differences in terms of performance for the various splitting criteria presented in \Cref{sec:methodology}. These splitting criteria include Gini, OGini, IG, WIG, and RI, which are utilised to build a DT classifier. Note that all other hyperparameters of the DT classifier remain fixed, with each DT classifier being named after the splitting criterion used during its training phase.

To conduct a comprehensive study, we evaluate each splitting criterion using a set of $45$ ordinal datasets collected from different OC works. Specifically, $41$ of them were obtained from \cite{27_gutierrez2015ordinal}, which is the the most extensive experimental study conducted to date in the OC field. The remaining $4$ datasets are well-established and frequently utilised \cite{48_janitza2016random}. A detailed overview of these datasets is provided in Appendix C.

For consistency and fairness, each splitting criterion is evaluated on every dataset over $20$ independent runs per dataset, using different seeds. To ensure comparability, the same set of seeds is applied across all splitting criteria. In each run, the maximum depth of the trees is adjusted using a stratified $5$-fold crossvalidation procedure, optimised in terms of Mean Absolute Error (MAE), and selecting the best value from the set $\{3, 5, 8, 16\}$. Moreover, the tree-growing method for the DT classifier is restricted to the one used in CART method \cite{31_leo1984classification}. In the special case of the WIG criterion, the normalisation parameter $\propto$ is set to $1$.

For the evaluation, we consider three different evaluation metrics that take the ordinality of the target variable into account: MAE, Quadratic Weighted Kappa (QWK), and Ranked Probability Score (RPS). The MAE is a widely utilised metric in OC. It measures the mean absolute difference between the observed and the predicted classes, considering the scores assigned by the function $v$. It is expressed as:
\begin{equation}
    \text{MAE} = \frac{1}{N} \sum_{i=1}^N | v(y_i) - \hat{y}_i|.
\end{equation}

The QWK is based on the kappa index \cite{vargas2023exponential}, which measures the inter-rater agreement on classifying elements into a set of categories. The value of the QWK ranges from $-1$ to $1$, with $1$ being the best possible score. This metric incorporates a penalisation term that grows quadratically with the distance between the two predicted classes. In this way, the QWK is expressed as:
\begin{equation}
    \text{QWK} = 1 - \frac{\sum_{q,j}W_{q,j}O_{q,j}}{\sum_{q,j}W_{q,j}E_{q,j}},
\end{equation}
where $\mathbf{W}$, $\mathbf{O}$ and $\mathbf{E}$ are matrices of size $Q \times Q$. $\mathbf{O}$ is the confusion matrix. $\mathbf{E}$ represents the expectation to obtain the observed agreement by random chance, where each element $E_{q,j}$ is computed as $E_{q,j} = \frac{\sum_i\mathds{1}(y_i=C_q)\sum_i\mathds{1}(\hat{y}_i=v(C_j))}{N}$. Here, $\hat{y}_i$ is the predicted category for the $i$-th instance. The value of the $W_{q,j}$ element of the quadratic weights matrix $W$ is computed as:
\begin{equation}
    W_{q,j} = \frac{(v(C_q)-v(C_j))^2}{(Q - 1)^2}.
\end{equation}

Finally, the RPS metric was introduced in \cite{epstein1969scoring} as a generalisation of the Brier score. It is a minimisation metric that measures the averaged difference between the predicted and the observed cumulative distribution of the ordinal classes:
\begin{equation}
    \text{RPS} = \frac{1}{N} \sum_{i=1}^N \sum_{q=1}^Q (\hat{\pi}_i(q) - \mathds{1}(y_i \leq C_q))^2,
\end{equation}
where $\hat{\pi}_i(q)$ denotes the estimated cumulative probability of observing class $C_q$ in the $i$-th sample, and $\mathds{1}(y_i \leq C_q) = 1$ when $y_i \leq C_q$ is true, and $0$ otherwise.

\section{Results} \label{sec:results}

The results obtained for the $20$ executions of each DT classifier using different splitting criteria are presented in \Cref{table:mean_results}, expressed as $\text{Mean}_{\text{STD}}$, where STD stands for the standard deviation. It can be observed that the ordinal splitting criteria, i.e. OGini, WIG, and RI, obtain better results for MAE and QWK than the nominal ones (Gini and IG). In terms of RPS, OGini and RI again outperform IG and Gini. However, in this case WIG fails to outperform the nominal metrics. It is important to remark that OGini obtains the best mean results for the three metrics, while the second best results are obtained by the WIG in MAE and QWK, and by the RI in terms of RPS.

\begin{table}[ht!]
\centering
\caption{Mean and standard deviation (STD) results (Mean$_{\text{STD}}$) for each method in terms of MAE, QWK and RPS. All datasets and splits are considered.}

\begin{tabular}{cccc}
\toprule \toprule
    Method  & MAE  ($\downarrow$)  & QWK ($\uparrow$) & RPS ($\downarrow$) \\
    \midrule
    Gini    & $0.875_{0.098}$   & $0.642_{0.071}$ & $0.745_{0.125}$ \\
    IG      & $0.880_{0.103}$   & $0.641_{0.073}$ & $0.749_{0.139}$ \\
    \midrule
    OGini   & $\mathbf{0.849_{0.094}}$ & $\mathbf{0.658_{0.068}}$ & $\mathbf{0.718_{0.129}}$ \\
    WIG     & $\mathit{0.869_{0.092}}$ & $\mathit{0.650_{0.067}}$ & $0.770_{0.123}$ \\
    RI      & $\mathit{0.869_{0.096}}$ & $0.646_{0.068}$ & $\mathit{0.719_{0.146}}$ \\
\bottomrule \bottomrule
\multicolumn{4}{l}{The best result is highlighted in \textbf{bold}, whereas the}\\
\multicolumn{4}{l}{second best is in \textit{italics}.}
\end{tabular}
\label{table:mean_results}
\end{table}

To analyse the statistical significance of the differences among the considered splitting criteria for MAE, QWK, and RPS performance metrics, the following statistical setup is used: 1) the non-parametric Kolmogorov-Smirnov normality test (K-S test) is firstly applied to verify if the results obtained for each metric follow a normal distribution; 2) given that the null hypothesis of normality is  accepted by K-S test in our experiments, an ANalysis Of the VAriance (ANOVA) test \cite{fisher1925theory, miller1997beyond} is performed. Specifically, an ANOVA II is applied to analyse the effects of two qualitative factors (the dataset and the splitting criterion) on the quantitative response, i.e. the values obtained in terms of MAE, QWK, and RPS metrics; 3) if the null hypothesis of the ANOVA II test, indicating no significant differences between means, is rejected, a Tukey post-hoc test is finally conducted to identify the specific groups between which significant differences exist. A significance level of $\alpha = 0.050$ is used for these statistical tests.

For further analysis on the scalability of these approaches, and following the previous statistical setup, two groups, based on the number of classes, $Q$, are considered. The motivation behind this lies in the assumption that the ordinal characteristics of a problem become more significant as $Q$ increases. As a result, the differences in performance between nominal and ordinal splitting methods are expected to be more pronounced in datasets with higher values of $Q$. Specifically, we first examine the datasets with $6$ or more classes ($Q \geq 6$) and then those with fewer than $6$ classes ($Q < 6$). This threshold of $6$ is chosen as it approximately balances the groups: the first group consists of $17$ datasets, while the second group includes $28$ datasets. 

\subsection{Statistical analysis for datasets with $Q \geq 6$}

In terms of MAE, the null hypothesis of normality of the K-S test is accepted for each of the $85$ samples of size $20$ ($17$ databases $\times \ 5$ splitting criteria). Subsequently, the ANOVA II test is performed to test whether the methodology and dataset (factors) have any significant impact on the mean values obtained in each metric. The results indicate the existence of significant differences between the mean results obtained by the different methodologies (p-value $< 0.001$). In view of these results, the Tukey post-hoc test is conducted. The results for this test are presented in \Cref{table:tukey_mae}, where the methods (splitting criteria) are grouped based on the statistical differences in their performance. Methods within the same group exhibit no significant differences, while those in different groups show statistically significant differences. As can be observed in the left part of  \Cref{table:tukey_mae} ($Q \geq 6$), group $1$ contains the best methodologies, OGini and RI, with no significant differences between them (p-value $=0.063$), and the group $3$ contains the worst methodologies, Gini and IG. This table also indicates significant differences between the pairs OGini-Gini, OGini-IG, WIG-IG, and RI-IG, always in favour of the ordinal methods. However, although RI presents a lower MAE, the difference against Gini is not significant (group $2$). Additionally, OGini stands as the best performing splitting criterion, obtaining significant differences against all the other criteria except for the RI.

\begin{table}[H]
\color{black}
    \centering
    \caption{\rev{Results of the post-hoc Tukey test for datasets with $Q \geq 6$ (left part) and for datasets with $Q < 6$ (right part), in terms of MAE.}}
    \setlength{\tabcolsep}{3.5pt}
    \begin{tabular}{lcccclcc}
        \toprule \toprule
        \multicolumn{8}{c}{Group Rank MAE ($\downarrow$)}\\
        \midrule
        \multicolumn{4}{c}{$Q \geq 6$} & & \multicolumn{3}{c}{$Q < 6$}\\
        \cmidrule{1-4} \cmidrule{6-8}
        Method          & 1         & 2     & 3     & & Method          & 1         & 2    \\
        \cmidrule{1-4} \cmidrule{6-8}
        OGini   & $1.393$  &        &       & \hspace{8pt} & OGini   & $0.519$  &        \\  
        RI      & $1.426$  & $1.426$  &       & \hspace{8pt} & Gini      & $0.526$  & $0.526$  \\
        WIG     &        & $1.429$  &       & \hspace{8pt} & IG     & & $0.529$  \\  
        Gini    &        & $1.454$  & $1.454$ & \hspace{8pt} & RI    & & $0.531$  \\
        IG      &        &        & $1.464$ & \hspace{8pt} & WIG      &        & $0.534$   \\
        \cmidrule{1-4} \cmidrule{6-8}
        p-value & $0.063$  & $0.153$  & $0.945$ & \hspace{8pt} & p-value & $0.161$  & $0.577$ \\ 
        \bottomrule \bottomrule
    \end{tabular}
    \label{table:tukey_mae}
\end{table}

For the QWK metric, the K-S test accepts the null hypothesis of normality, so the same procedure utilised for the MAE is conducted. The ANOVA II shows an overall statistically significant difference between the mean results across the splitting criteria (p-value $< 0.001$). Therefore, the post-hoc Tukey test is conducted (\Cref{table:tukey_qwk}). OGini and RI outperform both IG and Gini, and WIG surpasses IG. However, WIG shows no significant differences compared to Gini (p-value $= 0.089$). Similarly as for the MAE metric, OGini stands as the best approach, having no significant differences against RI and WIG (p-value $= 0.996$).

\begin{table}[H]
\color{black}
    \centering
    \caption{\rev{Results of the post-hoc Tukey test for datasets with $Q \geq 6$ (left part) and for datasets with $Q < 6$ (right part), in terms of QWK.}}
    \setlength{\tabcolsep}{3.5pt}
    \begin{tabular}{lcccclc}
        \toprule \toprule
        \multicolumn{7}{c}{Group Rank QWK ($\uparrow$)}\\
        \midrule
        \multicolumn{4}{c}{$Q \geq 6$} & & \multicolumn{2}{c}{$Q < 6$}\\
        \cmidrule{1-4} \cmidrule{6-7}
        Method          & 1         & 2     & 3     & & Method          & 1        \\
        \cmidrule{1-4} \cmidrule{6-7}
        OGini   & $0.665$ &        &        & \hspace{8pt} & OGini & $0.654$\\
        RI      & $0.656$ &        &        & \hspace{8pt} & WIG & $0.646$\\
        WIG     & $0.653$ & $0.653$ &        & \hspace{8pt} & RI & $0.643$\\
        Gini    &        & $0.641$ & $0.641$ & \hspace{8pt} & IG & $0.643$\\
        IG      &        &        & $0.639$ & \hspace{8pt} & Gini & $0.642$\\ 
        \cmidrule{1-4} \cmidrule{6-7}
        p-value & $0.996$  & $0.089$  & $0.143$ & \hspace{8pt} & p-value & $0.464$\\
        \bottomrule \bottomrule
    \end{tabular}
    \label{table:tukey_qwk}
\end{table}

Finally, in terms of RPS, once the hypothesis of normality is accepted, the ANOVA II test demonstrate a significant difference between the mean results of the five splitting criteria (p-value $ < 0.001$). The Tukey test (see \Cref{table:tukey_rps}) indicates that there are no significant differences in mean RPS between RI, OGini, Gini and IG (p-value $ = 0.082$). On the contrary, WIG is observed to be significantly worse than RI, Ogini and Gini. Additionally, we can observe that RI obtains the best results.

\begin{table}[H]
\color{black}
    \centering
    \caption{\rev{Results of the post-hoc Tukey test for datasets with $Q \geq 6$ (left part) and for datasets with $Q < 6$ (right part), in terms of RPS.}}
    \setlength{\tabcolsep}{3.5pt}
    \begin{tabular}{lccclcc}
        \toprule \toprule
        \multicolumn{7}{c}{Group Rank RPS ($\downarrow$)}\\
        \midrule
        \multicolumn{3}{c}{$Q \geq 6$} & & \multicolumn{3}{c}{$Q < 6$}\\
        \cmidrule{1-3} \cmidrule{5-7}
        Method          & 1         & 2     & & Method          & 1         & 2    \\
        \cmidrule{1-3} \cmidrule{5-7}
        RI      & $1.189$ &        & \hspace{8pt} & OGini      & $0.437$ &        \\
        OGini   & $1.190$ &        & \hspace{8pt} & Gini   & $0.440$ &        \\
        Gini    & $1.229$ &        & \hspace{8pt} & RI    & $0.442$ &        \\
        IG      & $1.236$ & $1.236$ & \hspace{8pt} & IG      & $0.446$ & $0.446$ \\
        WIG     &        & $1.284$ & \hspace{8pt} & WIG     &        & $0.463$ \\
        \cmidrule{1-3} \cmidrule{5-7}
        p-value & $0.082$  & $0.066$ & \hspace{8pt} & p-value & $0.590$  & $0.067$ \\  
        \bottomrule \bottomrule
    \end{tabular}
    \label{table:tukey_rps}
\end{table}

As a general comment, by comparing left and right parts of \Cref{table:tukey_mae,table:tukey_qwk,table:tukey_rps}, it is clear that differences are more evident in performance when the number of classes increases. This comes from the fact that the possible costs of misclassification increases, and ordinal splitting criteria can better exploit the ordinality of the target label and its relationship with the input data.

\subsection{Statistical analysis for datasets with $Q < 6$}

Focusing on the MAE metric, the hypothesis of normality is accepted, then, the ANOVA II is applied, showing that there are significant differences between the mean values of the five splitting criteria (p-value$=0.049$). Thus, the post-hoc Tukey test is conducted. The results are shown in the right part of \Cref{table:tukey_mae}. It can be observed that OGini achieves the best mean results. However, there are no significant differences between OGini, RI, and WIG with respect to Gini and IG.

In terms of QWK, since the hypothesis of normality is accepted, the ANOVA II test is applied indicating that no significant differences are found with respect to the mean results obtained by the splitting criteria, achieving a p-value$= 0.446$. In this case, the post-hoc Tukey test is not performed (in \Cref{table:tukey_qwk} all the splitting criteria are in the same group).

Finally, for the RPS metric the hypothesis of normality is accepted again. Then, the ANOVA II test is conducted, showing significant differences in the mean values between the different splitting criteria. Hence, the post-hoc Tukey test is performed. The results (see \Cref{table:tukey_rps}) indicate that OGini is the best splitting criteria in mean results, but no significant differences are found with respect to Gini and IG.

\subsection{\rev{Gini vs OGini: in-depth analysis of a real case}}

\begin{figure*}[ht!]
    \centering
    \begin{subfigure}{0.49\textwidth}
        \includegraphics[width=\textwidth]{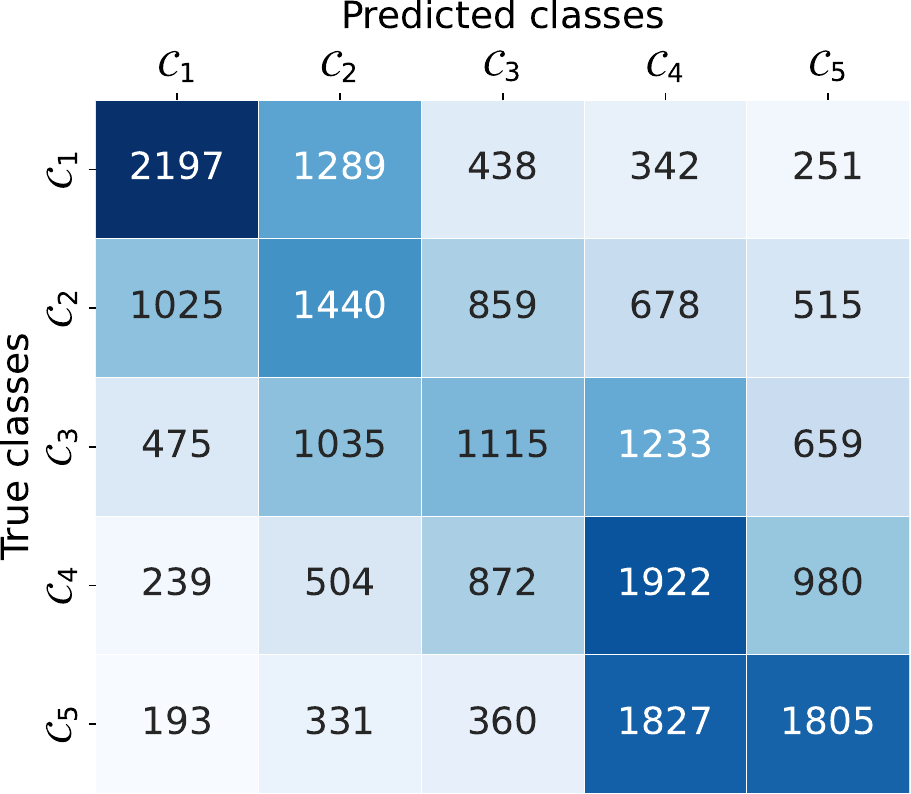}
        \subcaption{Gini}
    \end{subfigure}
    \hfill
    \begin{subfigure}{0.49\textwidth}
        \includegraphics[width=\textwidth]{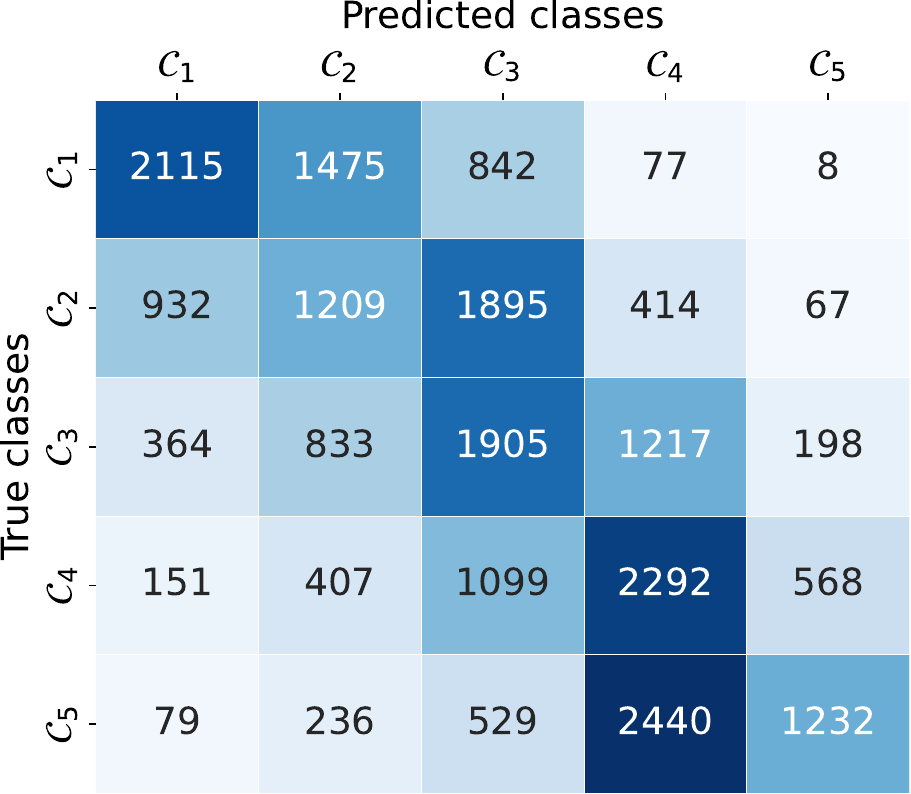}
        \subcaption{OGini}
    \end{subfigure}
    \caption{\rev{Confusion matrices obtained in the test set by Gini and OGini in the \texttt{census-2-5} dataset.}}
    \label{fig:confusion_matrices}
\end{figure*}

In this subsection, the benefit of using ordinal splitting criteria compared to their nominal counterparts is illustrated in \Cref{fig:confusion_matrices} for a specific real case (\texttt{census2-5} dataset). For this purpose, two DT classifiers are trained using the Gini and OGini splitting criteria, respectively, with the same random seed $0$. The resulting confusion matrices for each classifier are presented below, highlighting the differences in how the models classify the data. These matrices demonstrate the impact of considering ordinal information in the splitting process, as the OGini splitting criterion better accounts for the ordinal relationships between classes, leading to closer targets in the ordinal scale when compared to the Gini splitting criterion.



It can be observed that the use of the OGini splitting criterion significantly reduces errors in the extreme classes compared to the Gini criterion. For example, the number of instances where patterns from class $C_1$ are misclassified as class $C_5$ decreases dramatically from $251$ to $8$. Similarly, the number of instances where patterns from class $C_5$ are misclassified as class $C_1$ is reduced from $193$ to $79$. Moreover, considering the case of class $C_3$, which lies in the middle of the ordinal scale, the number of patterns misclassified as class $C_1$ and class $C_5$ is reduced from $475$ to $364$, and from $659$ to $198$, respectively. These improvements in the confusion matrices are also evident in the performance metrics, which are presented in \Cref{tab:ogini_vs_gini_results}.

\begin{table}[]
    \centering
    \begin{tabular}{rccc}
    \toprule\toprule
    Splitting criterion   & MAE ($\downarrow$) & QWK ($\uparrow$) & RPS ($\downarrow$) \\
    \midrule
    Gini        & 0.948     & 0.534     & 0.948 \\
    OGini       & 0.793     & 0.648     & 0.571 \\
    \bottomrule\bottomrule
    \end{tabular}
    \caption{\rev{MAE, QWK and RPS obtained in the test set by Gini and OGini in the \texttt{census-2-5} dataset.}}
    \label{tab:ogini_vs_gini_results}
\end{table}

These improvements highlight the ability of ordinal splitting criteria to better respect the ordinal relationships between classes, particularly in cases where the misclassification involves distant classes on the ordinal scale.

\section{Conclusions}\label{sec:conclusions}

This work provides an exhaustive survey of the ordinal splitting criteria proposed in the literature: Ordinal Gini (OGini), Weighted Information Gain (WIG), and Ranking Impurity (RI), where the first two are ordinal adaptations of the well-known Gini and Information Gain (IG). A unified notation has been established to clearly define the different splitting criteria, as well as the impurity measures associated to them: Gini-index and Shannon-entropy for the nominal splitting criteria, and OGini-index, weighted entropy, and RI, for the ordinal ones.

The methodologies have been empirically evaluated using a wide range of $45$ ordinal problems in terms of three different ordinal metrics: Mean Absolute Error (MAE), Quadratic Weighted Kappa (QWK), and Ranked Probability Score (RPS). The results presented in this work confirm the superiority of the ordinal splitting criteria over their nominal counterparts, in the mean values obtained for MAE, QWK and RPS, with OGini standing as the best approach. \rev{Furthermore, a subsequent study statistically analysing the results in datasets with $6$ or more classes, where the ordinality of the output variable is more pronounced, proves significant differences in favour of the ordinal approaches. To this end, a two-way ANOVA was applied to assess the influence of both the splitting criterion and the dataset on performance outcomes. To further identify specific group differences, Tukey’s post-hoc test was employed. The results demonstrate statistically significant differences in favour of the ordinal approaches, particularly highlighting that OGini consistently and significantly outperforms both Gini and IG in terms of MAE and QWK.}

\rev{To promote transparency and reproducibility, the $45$ OC datasets, all the individual results, the complete source code, the implementation of the splitting criteria considered in this study, and detailed instructions for reproducing all the experiments have been made publicly available through the associated website\footnote{\url{https://www.uco.es/grupos/ayrna/ordinal-trees}}.}

\section{\rev{Limitations and Future Work}} \label{sec:future}
\rev{Traditional decision tree algorithms, such as CART or C4.5, treat categorical features in a nominal manner, disregarding the inherent ordering in ordinal variables. This nominal approach is suboptimal in classification tasks involving an ordinal target, as it fails to exploit the ordinal structure present in the data. Furthermore, nominal splits arbitrarily partition the set of categories without considering their rank order, leading to splits that may not reflect the true underlying decision boundaries.

In contrast, ordinal splitting strategies incorporate the natural order of the categories into the tree growing phase. This typically results in more semantically meaningful splits, improved alignment with the true class structure, and, consequently, enhanced predictive performance, particularly when the evaluation metric is sensitive to the ordinal structure (e.g., mean absolute error, ranked probability score).

Beyond predictive accuracy, ordinal-aware trees support more stable and robust models, particularly when applied to small or imbalanced datasets, where preserving the order of the classes can act as an implicit regularization mechanism. They also serve as a natural foundation for ensemble methods in ordinal classification settings, such as ordinal random forests or ordinal boosting.}

We recognise the potential for further development of effective ordinal splitting criteria. In this sense, we also provide a framework that eases the testing and validation of new DT methods and splitting criteria in a plug-and-play way. Hence, the unified mathematical notation, code, and experimental study presented in this work offer additional opportunities to explore this field, serving as a baseline for future developments.

\rev{The development of context-appropriate splitting criteria can improve performance in many real-world applications, given that tree-based techniques are widely utilised and have proven to be one the best performing strategies within machine learning.} A potential future line of research is to study the application of these splitting criteria in ensembles of trees, where the ordinality can be further exploited employing ordinal ensemble techniques such as the ordinal forest proposed in \cite{hornung2020ordinal}. We also would like to extend the ordinal problems archive used for comparison, as this will facilitate the validation of new proposals. We greatly appreciate any contributions to this archive.

\section*{Acknowledgments}
The present study has been supported by the ``Agencia Estatal de Investigación (España)'' (grant ref.: PID2023-150663NB-C22 / AEI / 10.13039 / 501100011033), by the European Commission, AgriFoodTEF (grant ref.: DIGITAL-2022-CLOUD-AI-02, 101100622), by the Secretary of State for Digitalization and Artificial Intelligence ENIA International Chair (grant ref.: TSI-100921-2023-3), and by the University of Córdoba and Junta de Andalucía (grant ref.: PP2F\_L1\_15). R. Ayllón-Gavilán has been supported by the ``Instituto de Salud Carlos III'' (ISCIII) and EU (grant ref.: FI23/00163).

\section*{Declaration of competing interest}
The authors declare that they have no known competing financial interests or personal relationships that could have appeared to influence the work reported in this paper.

\section*{Author contributions: CRediT}
\textbf{R. Ayllón-Gavilán}: Data curation, Formal analysis, Investigation, Methodology, Software, Validation, Writing - Original draft. \textbf{F. J. Martínez-Estudillo}: Conceptualization, Formal analysis, Investigation, Methodology, Writing - Original draft, Writing - review and editing. \textbf{D. Guijo-Rubio}: Resources, Validation, Visualization, Writing - review and editing. \textbf{C. Hervás-Martínez}: Conceptualization, Formal analysis, Resources, Supervision, Writing - review and editing. \textbf{P. A. Gutiérrez}: Data curation, Funding acquisition, Project administration, Supervision, Writing - review and editing.

\bibliographystyle{elsarticle-num}
\bibliography{bibliography_dois}

\end{document}


\begin{frontmatter}

\title{Splitting criteria for ordinal decision trees: an experimental study}

\author[imibic,pduco]{Rafael Ayllón-Gavilán}
\author[uloyola]{Francisco José Martínez-Estudillo}
\author[uco]{David Guijo-Rubio\corref{cor}}\cortext[cor]{Department of Computer Science and Numerical Analysis, Universidad de Córdoba, Campus de Rabanales, Ctra. N-IVa, Km. 396, Córdoba, 14071.} \ead{dguijo@uco.es}
\author[uco]{César Hervás-Martínez}
\author[uco]{Pedro A. Gutiérrez}

\affiliation[imibic]{organization={Department of Clinical-Epidemiological Research in Primary Care, IMIBIC},
            addressline={Avda. Menéndez Pidal S/N},
            city={Córdoba},
            postcode={14004},
            country={Spain}}

\affiliation[pduco]{organization={Programa de doctorado en Computación Avanzada, Energía y Plasmas, Universidad de Córdoba},
            addressline={Campus de Rabanales, Ctra. N-IVa, Km. 396},
            city={Córdoba},
            postcode={14071},
            country={Spain}}
            
\affiliation[uloyola]{organization={Department of Quantitative Methods, Universidad Loyola Andalucía},
            addressline={C. Escritor Castilla Aguayo, 4},
            city={Córdoba},
            postcode={14004},
            country={Spain}}

\affiliation[uco]{organization={Departamento de Ciencia de la Computación e Inteligencia Artificial, Universidad de Córdoba},
            addressline={Campus de Rabanales, Ctra. N-IVa, Km. 396},
            city={Córdoba},
            postcode={14071},
            country={Spain}}

\begin{abstract}
Ordinal Classification (OC) addresses those classification tasks where the labels exhibit a natural order. Unlike nominal classification, which treats all classes as \rev{mutually exclusive and unordered}, OC takes the ordinal relationship into account, producing more accurate and relevant results. This is particularly critical in applications where the magnitude of classification errors \rev{has significant consequences}. Despite this, OC problems are often tackled using nominal methods, leading to suboptimal solutions. Although decision trees are \rev{among} the most popular classification approaches, ordinal tree-based approaches have received less attention when compared to other classifiers. This work provides a comprehensive survey of ordinal splitting criteria, standardising the notations used in the literature \rev{to enhance clarity and consistency}. Three ordinal splitting criteria, Ordinal Gini (OGini), Weighted Information Gain (WIG), and Ranking Impurity (RI), are compared to the nominal counterparts of the first two (Gini and information gain), by incorporating them into a decision tree classifier. An extensive repository considering $45$ publicly available OC datasets is presented, supporting the first experimental comparison of ordinal and nominal splitting criteria using well-known OC evaluation metrics. \rev{The results have been statistically analysed, highlighting that OGini stands out as the best ordinal splitting criterion to date, reducing the mean absolute error achieved by Gini by more than $3.02\%$.} \rev{To promote reproducibility, all source code developed, a detailed guide for reproducing the results, the $45$ OC datasets, and the individual results for all the evaluated methodologies are provided.}
\end{abstract}

\begin{keyword}
Ordinal classification \sep ordinal regression \sep ordinal trees \sep impurity measures \sep splitting criteria \sep ordinal Gini \sep ordinal information gain \sep ranking impurity
\end{keyword}
\end{frontmatter}

\begin{appendices}

\section{Datasets characteristics}
The main characteristics of the datasets considered in the experiments are presented in \Cref{table:dataset_info}. We consider a total of $45$ databases of which $21$ are original ordinal classification problems, and the remaining $24$ are discretised regression datasets. It can be observed that the largest training set comprises a total of $3499$ in the \textit{nhanes} dataset, and the lowest a total of $18$ patterns (\textit{contact-lenses}). With respect to the imbalance ratio, note that the discretised regression datasets present a perfect balance due to the equal-frequency discretisation strategy employed. Datasets are available in the associated website\footnote{\url{https://www.uco.es/grupos/ayrna/ordinal-trees}}.

\begin{table}[ht!]
\setlength{\tabcolsep}{5.5pt}
\renewcommand{\arraystretch}{1.15}
\centering
\caption{Characteristics of the datasets used in this work. \#Train and \#Test stand for the number of training and testing patterns, respectively, $Q$ indicates the number of classes, and $K$ is the number of input features. The IR columns represents the Imbalance Ratio, computed as the number of patterns of the most represented class divided by the number of patterns of the less represented class.}
\label{table:dataset_info}
\resizebox{\textwidth}{!}{
\begin{tabular}{rccccc|rccccc}
\toprule \toprule
\multicolumn{6}{c|}{\textbf{Original OC problems}} & \multicolumn{6}{c}{\textbf{Discretised regression problems}} \\
\midrule
Dataset & \#Train & \#Test & $Q$ & $K$ & IR & Dataset & \#Train & \#Test & $Q$ & $K$ & IR\\
\midrule
ERA     & $750$     & $250$     & $9$       & $4$       & $9.714$  & abalone-10       & $1000$        & $3177$    & $10$      & $10$      & $1.000$ \\
ESL     & $366$     & $122$     & $9$       & $4$       & $50.500$  & bank1-10     & $50$      & $8142$    & $10$      & $8$       & $1.000$ \\
LEV     & $750$     & $250$     & $5$       & $4$       & $15.100$  & bank1-5      & $50$      & $8142$    & $5$       & $8$       & $1.000$ \\
SWD     & $750$     & $250$     & $4$       & $10$      & $12.458$  & bank2-10     & $75$      & $8117$    & $10$      & $32$      & $1.143$ \\
automobile      & $153$     & $52$  & $6$       & $71$      & $25.500$  & bank2-5      & $75$      & $8117$    & $5$       & $32$      & $1.000$ \\
balancescale        & $468$     & $157$     & $3$       & $4$       & $6.000$  & calhousing-10        & $150$     & $20490$   & $10$      & $8$       & $1.000$ \\
bondrate        & $42$      & $15$  & $4$       & $37$      & $6.000$  & calhousing-5     & $150$     & $20490$   & $5$       & $8$       & $1.000$ \\
car     & $1296$        & $432$     & $4$       & $21$      & $18.510$  & census1-10       & $175$     & $22609$   & $10$      & $8$       & $1.059$ \\
contact-lenses       & $18$      & $6$   & $3$       & $6$       & $3.667$  & census1-5        & $175$     & $22609$   & $5$       & $8$       & $1.000$ \\
eucalyptus      & $552$     & $184$     & $5$       & $91$      & $2.064$  & census2-10       & $200$     & $22584$   & $10$      & $16$      & $1.000$ \\
mammoexp        & $276$     & $136$     & $3$       & $5$       & $3.204$  & census2-5        & $200$     & $22584$   & $5$       & $16$      & $1.000$ \\
newthyroid      & $161$     & $54$  & $3$       & $5$       & $4.870$  & computer1-10     & $100$     & $8092$    & $10$      & $12$      & $1.000$ \\
nhanes      & $3499$        & $1724$    & $5$       & $30$      & $10.669$  & computer1-5      & $100$     & $8092$    & $5$       & $12$      & $1.000$ \\
pasture     & $27$      & $9$   & $3$       & $25$      & $1.000$  & computer2-10     & $125$     & $8067$    & $10$      & $21$      & $1.083$ \\
squash-stored        & $39$      & $13$  & $3$       & $51$      & $3.000$  & computer2-5      & $125$     & $8067$    & $5$       & $21$      & $1.000$ \\
squash-unstored      & $39$      & $13$  & $3$       & $52$      & $6.000$  & housing     & $300$     & $206$     & $5$       & $13$      & $1.000$ \\
support     & $489$     & $241$     & $5$       & $19$      & $39.200$  & housing-10       & $300$     & $206$     & $10$      & $13$      & $1.000$ \\
tae     & $113$     & $38$  & $3$       & $54$      & $1.083$  & machine     & $150$     & $59$  & $5$       & $6$       & $1.000$ \\
toy     & $225$     & $75$  & $5$       & $2$       & $2.708$  & machine10       & $150$     & $59$  & $10$      & $6$       & $1.000$ \\
vlbw        & $115$     & $57$  & $9$       & $19$      & $3.143$  & pyrim       & $50$      & $24$  & $5$       & $26$      & $1.000$ \\
winequality-red      & $1199$        & $400$     & $6$       & $11$      & $63.750$  & pyrim-10     & $50$      & $24$  & $10$      & $26$      & $1.000$ \\
                    &               &           &           &           &            & stock       & $600$     & $350$     & $5$       & $9$       & $1.000$ \\
                    &               &           &           &           &            & stock-10     & $600$     & $350$     & $10$      & $9$       & $1.000$ \\
                    &               &           &           &           &            & abalone-5   & $1000$        & $3177$    & $10$      & $5$      & $1.000$ \\
\bottomrule \bottomrule
\end{tabular}
}
\end{table}

\section{Runtime and RAM consumption}
In \Cref{tab:runtime}, we present the mean runtime of each technique along the 45 datasets. It can be observed that all methods present a very similar runtime. This is attributed to the fact that each method is ultimately the same decision tree, where only the splitting method changes, which generally does not have a significant impact on performance. Similarly, with respect to the RAM usage, all methods consume approximately the same memory.

\begin{table}[]
    \setlength{\tabcolsep}{3.5pt}
    \renewcommand{\arraystretch}{1.15}
    \centering
    \caption{Mean runtime (in seconds) of each method along the $45$ datasets.}
    \begin{tabular}{rccccccc}
        \toprule\toprule
        \multirow{2.5}{*}{Splitting criterion}& \multicolumn{3}{c}{Time (s)} & & \multicolumn{3}{c}{RAM Usage (MB)} \\
        \cmidrule{2-4} \cmidrule{6-8}
              & Mean	    & Min  & Max  & & Mean    & Min  & Max \\
        \midrule
        IG			& $66.520$	        & $0.532$      & $242.943$     &   & $114.983$       & $113.254$      & $120.559$         \\ 
        Gini		& $69.928$	        & $0.512$      & $241.741$      &  & $115.091$       & $113.121$      & $121.453$        \\
        OGini		& $70.158$	        & $0.502$      & $241.001$      &  & $114.542$       & $112.099$      & $121.619$       \\
        WIG			& $71.098$	        & $0.587$      & $243.098$      &  & $114.890$       & $112.988$      & $119.201$    \\
        RI			& $69.903$	        & $0.546$      & $242.908$      &  & $114.500$       & $113.751$      & $120.009$     \\
        \bottomrule\bottomrule
    \end{tabular}
    \label{tab:runtime}
\end{table}

\section{Results per method, dataset and metric}
In \Cref{tab:results}, we provide the results of each method--dataset pair in terms of the three performance metrics: Mean Absolute Error (MAE), Quadratic Weighted Kappa (QWK) and Ranked Probability Score (RPS). Each method is trained on each dataset 20 times using different random seeds and shuffling the train-test stratified partitions, the values provided in \Cref{tab:results} are the averages of the results obtained across those 20 executions.

\begin{sidewaystable}[]
    \centering
    \caption{Average results for each method--dataset--metric combination. The whole set of experimental results are available on the associated webpage: \\ \url{https://www.uco.es/grupos/ayrna/ordinal-trees}.}
    \setlength{\tabcolsep}{5.5pt}
    \renewcommand{\arraystretch}{0.7}
    \begin{tabular}{cl|ccccc|ccccc|ccccc}
        \toprule\toprule
        &     \multirow{2.5}{*}{Dataset}    & \multicolumn{5}{c|}{MAE ($\downarrow$)} & \multicolumn{5}{c|}{QWK ($\uparrow$)} & \multicolumn{5}{c}{RPS ($\downarrow$)} \\
        \cmidrule{3-17}
        &  & IG & Gini & OGini & WIG & RI & IG & Gini & OGini & WIG & RI & IG & Gini & OGini & WIG & RI  \\
        \midrule
        \multirow{21}{*}{\rotatebox[origin=c]{90}{Original OC Problems}} & automobile & 0.280 & 0.306 & 0.333 & 0.327 & 0.316 & 0.848 & 0.833 & 0.825 & 0.815 & 0.836 & 0.280 & 0.305 & 0.329 & 0.325 & 0.316  \\
        & balance-scale & 0.273 & 0.269 & 0.261 & 0.260 & 0.278 & 0.805 & 0.807 & 0.817 & 0.818 & 0.799 & 0.270 & 0.264 & 0.258 & 0.258 & 0.277  \\
        & bondrate & 0.757 & 0.707 & 0.653 & 0.693 & 0.810 & 0.125 & 0.170 & 0.197 & 0.140 & 0.026 & 0.602 & 0.542 & 0.509 & 0.552 & 0.615  \\
        & car & 0.035 & 0.044 & 0.041 & 0.040 & 0.047 & 0.959 & 0.950 & 0.952 & 0.952 & 0.946 & 0.035 & 0.044 & 0.041 & 0.040 & 0.047  \\
        & contact-lenses & 0.383 & 0.383 & 0.350 & 0.383 & 0.367 & 0.389 & 0.389 & 0.471 & 0.389 & 0.431 & 0.348 & 0.348 & 0.333 & 0.348 & 0.341  \\
        & ERA & 1.384 & 1.348 & 1.357 & 1.334 & 1.367 & 0.537 & 0.537 & 0.533 & 0.525 & 0.521 & 0.889 & 0.891 & 0.882 & 0.900 & 0.888  \\
        & ESL & 0.950 & 0.940 & 0.948 & 0.939 & 0.945 & 0.722 & 0.721 & 0.718 & 0.727 & 0.719 & 0.883 & 0.870 & 0.883 & 0.879 & 0.898  \\
        & eucalyptus & 0.446 & 0.437 & 0.429 & 0.469 & 0.471 & 0.851 & 0.857 & 0.859 & 0.844 & 0.843 & 0.402 & 0.371 & 0.392 & 0.418 & 0.462  \\
        & LEV & 0.423 & 0.421 & 0.423 & 0.421 & 0.421 & 0.692 & 0.693 & 0.692 & 0.693 & 0.693 & 0.323 & 0.322 & 0.322 & 0.322 & 0.322  \\
        & mammoexp & 0.566 & 0.570 & 0.570 & 0.589 & 0.625 & 0.132 & 0.128 & 0.114 & 0.073 & 0.027 & 0.382 & 0.381 & 0.383 & 0.387 & 0.406  \\
        & newthyroid & 0.087 & 0.095 & 0.100 & 0.069 & 0.060 & 0.847 & 0.843 & 0.829 & 0.882 & 0.895 & 0.082 & 0.092 & 0.093 & 0.066 & 0.060  \\
        & nhanes & 0.713 & 0.712 & 0.719 & 0.732 & 0.715 & 0.205 & 0.204 & 0.208 & 0.276 & 0.237 & 0.501 & 0.502 & 0.502 & 0.504 & 0.501  \\
        & pasture & 0.228 & 0.217 & 0.222 & 0.250 & 0.328 & 0.834 & 0.839 & 0.836 & 0.814 & 0.733 & 0.214 & 0.206 & 0.211 & 0.224 & 0.329  \\
        & squash-stored & 0.431 & 0.438 & 0.423 & 0.412 & 0.400 & 0.479 & 0.446 & 0.478 & 0.525 & 0.514 & 0.408 & 0.428 & 0.398 & 0.387 & 0.401  \\
        & squash-unstored & 0.242 & 0.242 & 0.242 & 0.227 & 0.246 & 0.720 & 0.721 & 0.721 & 0.714 & 0.712 & 0.243 & 0.239 & 0.239 & 0.213 & 0.242  \\
        & support & 0.241 & 0.240 & 0.259 & 0.287 & 0.268 & 0.943 & 0.943 & 0.941 & 0.934 & 0.942 & 0.192 & 0.190 & 0.194 & 0.210 & 0.198  \\
        & SWD & 0.473 & 0.469 & 0.467 & 0.480 & 0.494 & 0.535 & 0.540 & 0.542 & 0.539 & 0.526 & 0.332 & 0.328 & 0.328 & 0.342 & 0.334  \\
        & tae & 0.571 & 0.572 & 0.607 & 0.575 & 0.600 & 0.385 & 0.399 & 0.370 & 0.414 & 0.310 & 0.532 & 0.531 & 0.532 & 0.505 & 0.583  \\
        & toy & 0.123 & 0.127 & 0.137 & 0.127 & 0.111 & 0.954 & 0.953 & 0.948 & 0.952 & 0.959 & 0.122 & 0.126 & 0.136 & 0.125 & 0.111  \\
        & vlbw & 2.615 & 2.585 & 2.463 & 2.470 & 2.477 & 0.186 & 0.187 & 0.281 & 0.240 & 0.288 & 2.312 & 2.106 & 2.001 & 2.164 & 1.825  \\
        & winequality-red & 0.482 & 0.480 & 0.480 & 0.476 & 0.475 & 0.472 & 0.467 & 0.473 & 0.498 & 0.461 & 0.412 & 0.376 & 0.374 & 0.426 & 0.357  \\
        \midrule
        \multirow{24}{*}{\rotatebox[origin=c]{90}{Discretised Regression Problems}} & abalone & 0.856 & 0.832 & 0.834 & 0.847 & 0.852 & 0.610 & 0.612 & 0.625 & 0.624 & 0.630 & 0.636 & 0.585 & 0.612 & 0.603 & 0.577  \\
        & abalone10 & 1.806 & 1.746 & 1.750 & 1.814 & 1.855 & 0.628 & 0.642 & 0.645 & 0.638 & 0.637 & 1.224 & 1.226 & 1.246 & 1.218 & 1.289  \\
        & bank1-10 & 1.444 & 1.492 & 1.254 & 1.375 & 1.286 & 0.758 & 0.744 & 0.818 & 0.781 & 0.806 & 1.334 & 1.378 & 1.196 & 1.259 & 1.108  \\
        & bank1-5 & 0.675 & 0.668 & 0.635 & 0.674 & 0.613 & 0.747 & 0.752 & 0.776 & 0.747 & 0.777 & 0.561 & 0.631 & 0.570 & 0.605 & 0.546  \\
        & bank2-10 & 2.676 & 2.759 & 2.524 & 2.621 & 2.615 & 0.284 & 0.282 & 0.365 & 0.331 & 0.331 & 2.218 & 2.121 & 2.107 & 2.267 & 2.126  \\
        & bank2-5 & 1.268 & 1.250 & 1.190 & 1.239 & 1.249 & 0.313 & 0.309 & 0.373 & 0.340 & 0.326 & 1.052 & 1.084 & 1.043 & 1.153 & 1.100  \\
        & calhousing-10 & 1.945 & 1.928 & 1.849 & 1.991 & 1.896 & 0.573 & 0.578 & 0.610 & 0.561 & 0.597 & 1.760 & 1.555 & 1.578 & 1.889 & 1.733  \\
        & calhousing-5 & 0.895 & 0.890 & 0.892 & 0.929 & 0.910 & 0.571 & 0.576 & 0.574 & 0.561 & 0.577 & 0.761 & 0.758 & 0.766 & 0.858 & 0.721  \\
        & census1-10 & 1.837 & 1.849 & 1.787 & 1.817 & 1.784 & 0.629 & 0.615 & 0.641 & 0.630 & 0.645 & 1.579 & 1.611 & 1.554 & 1.738 & 1.484  \\
        & census1-5 & 0.861 & 0.839 & 0.855 & 0.882 & 0.842 & 0.611 & 0.624 & 0.625 & 0.612 & 0.627 & 0.727 & 0.749 & 0.722 & 0.791 & 0.666  \\
        & census2-10 & 2.002 & 1.935 & 1.890 & 1.952 & 1.918 & 0.561 & 0.580 & 0.605 & 0.582 & 0.593 & 1.680 & 1.855 & 1.674 & 1.803 & 1.606  \\
        & census2-5 & 0.915 & 0.904 & 0.887 & 0.929 & 0.899 & 0.574 & 0.578 & 0.594 & 0.571 & 0.589 & 0.797 & 0.733 & 0.732 & 0.854 & 0.676  \\
        & computer1-10 & 1.524 & 1.456 & 1.433 & 1.460 & 1.455 & 0.716 & 0.738 & 0.743 & 0.736 & 0.738 & 1.348 & 1.346 & 1.327 & 1.413 & 1.278  \\
        & computer1-5 & 0.693 & 0.700 & 0.671 & 0.692 & 0.693 & 0.711 & 0.708 & 0.731 & 0.725 & 0.713 & 0.624 & 0.643 & 0.608 & 0.645 & 0.580  \\
        & computer2-10 & 1.372 & 1.283 & 1.292 & 1.355 & 1.355 & 0.754 & 0.787 & 0.788 & 0.768 & 0.768 & 1.220 & 1.191 & 1.054 & 1.303 & 1.192  \\
        & computer2-5 & 0.605 & 0.622 & 0.593 & 0.618 & 0.604 & 0.766 & 0.755 & 0.778 & 0.766 & 0.771 & 0.545 & 0.550 & 0.509 & 0.581 & 0.490  \\
        & housing & 0.498 & 0.500 & 0.503 & 0.527 & 0.503 & 0.830 & 0.820 & 0.830 & 0.812 & 0.828 & 0.397 & 0.414 & 0.407 & 0.475 & 0.365  \\
        & housing10 & 1.133 & 1.157 & 1.094 & 1.144 & 1.117 & 0.818 & 0.815 & 0.835 & 0.818 & 0.835 & 1.006 & 0.969 & 0.924 & 1.069 & 0.911  \\
        & machine & 0.531 & 0.538 & 0.508 & 0.519 & 0.519 & 0.810 & 0.807 & 0.821 & 0.826 & 0.808 & 0.439 & 0.497 & 0.454 & 0.494 & 0.435  \\
        & machine10 & 1.118 & 1.136 & 1.103 & 1.088 & 1.104 & 0.834 & 0.828 & 0.838 & 0.844 & 0.841 & 1.041 & 1.079 & 0.965 & 1.046 & 1.034  \\
        & pyrim & 0.850 & 0.885 & 0.858 & 0.835 & 0.831 & 0.620 & 0.595 & 0.614 & 0.615 & 0.644 & 0.786 & 0.821 & 0.766 & 0.825 & 0.749  \\
        & pyrim10 & 1.927 & 1.885 & 1.821 & 1.729 & 1.863 & 0.595 & 0.604 & 0.638 & 0.694 & 0.619 & 1.731 & 1.768 & 1.686 & 1.649 & 1.692  \\
        & stock & 0.158 & 0.166 & 0.156 & 0.168 & 0.156 & 0.958 & 0.956 & 0.958 & 0.955 & 0.959 & 0.152 & 0.162 & 0.152 & 0.163 & 0.154  \\
        & stock10 & 0.328 & 0.350 & 0.320 & 0.340 & 0.349 & 0.973 & 0.971 & 0.974 & 0.971 & 0.972 & 0.320 & 0.340 & 0.309 & 0.338 & 0.338  \\
        \bottomrule\bottomrule
    \end{tabular}
    
    \label{tab:results}
\end{sidewaystable}

\end{appendices}